\documentclass[runningheads]{llncs}
\usepackage{graphicx}
\usepackage{hyperref}
\usepackage{tikz}
\usepackage{comment}
\usepackage{amsmath,amssymb} 
\usepackage{color}
\usepackage{xspace}

\begin{document}
\providecommand{\MethodName}{CLASTER\xspace}
 \makeatletter\newcommand\myparagraph{\@startsection{paragraph}{4}{\z@}
  {.1em \@plus1ex \@minus.2ex}{-.5em}{\normalfont\normalsize\bfseries}}\makeatother
\pagestyle{headings}
\mainmatter
\def\ECCVSubNumber{3591}  

\title{CLASTER:\\Clustering with Reinforcement Learning for Zero-Shot Action Recognition} 


\titlerunning{CLASTER}
%
\author{Shreyank N Gowda\inst{1} \and
Laura Sevilla-Lara\inst{1}  \and \\
Frank Keller\inst{1} \and
Marcus Rohrbach\inst{2}}
\authorrunning{Shreyank N Gowda et al.}
%
\institute{University of Edinburgh \and
Meta AI\\
}

\maketitle

\begin{abstract}
Zero-Shot action recognition is the task of recognizing action classes without visual examples. 
The problem can be seen as learning a representation
on seen classes which generalizes well to instances of unseen classes, without losing discriminability 
between classes. Neural networks are able to model highly complex boundaries between visual classes, which explains their success as supervised models.
However, in Zero-Shot learning, these highly specialized class boundaries may overfit to the seen classes and not transfer well from seen to unseen classes. We propose a  novel cluster-based representation, which regularizes the learning process, yielding a representation that
generalizes well to instances from unseen classes. 
We optimize the clustering using reinforcement learning, which we observe is critical. We call the proposed method \MethodName and observe that it consistently outperforms the state-of-the-art in all standard Zero-Shot video datasets, including UCF101, HMDB51 and Olympic Sports; both in the standard Zero-Shot evaluation and the generalized Zero-Shot learning. We see improvements of up to 11.9\% over SOTA. 

Project Page: \url{https://sites.google.com/view/claster-zsl/home}
\end{abstract}

\section{Introduction}

Research on action recognition in videos has made rapid progress in the last years, with models becoming more accurate and even some datasets becoming saturated. Much of this progress has depended on large scale training sets. However, it is often not practical to collect thousands of video samples for a new class. This idea has led to research in the Zero-Shot learning (ZSL) domain, where training occurs in a set of seen classes, and testing occurs in a set of unseen classes. In particular, in the case of video ZSL, each class label is typically enriched with semantic embeddings. These embeddings are sometimes manually annotated, by providing attributes of the class, and other times computed automatically using language models of the class name or class description. 
At test time the semantic embedding of the predicted seen class is used to search for a nearest neighbor in the space of semantic embeddings of unseen classes. 
%
\\

While ZSL is potentially a very useful technology, this standard pipeline poses a fundamental representation challenge. 
\begin{figure}
    \centering
    \includegraphics[width=0.49\linewidth]{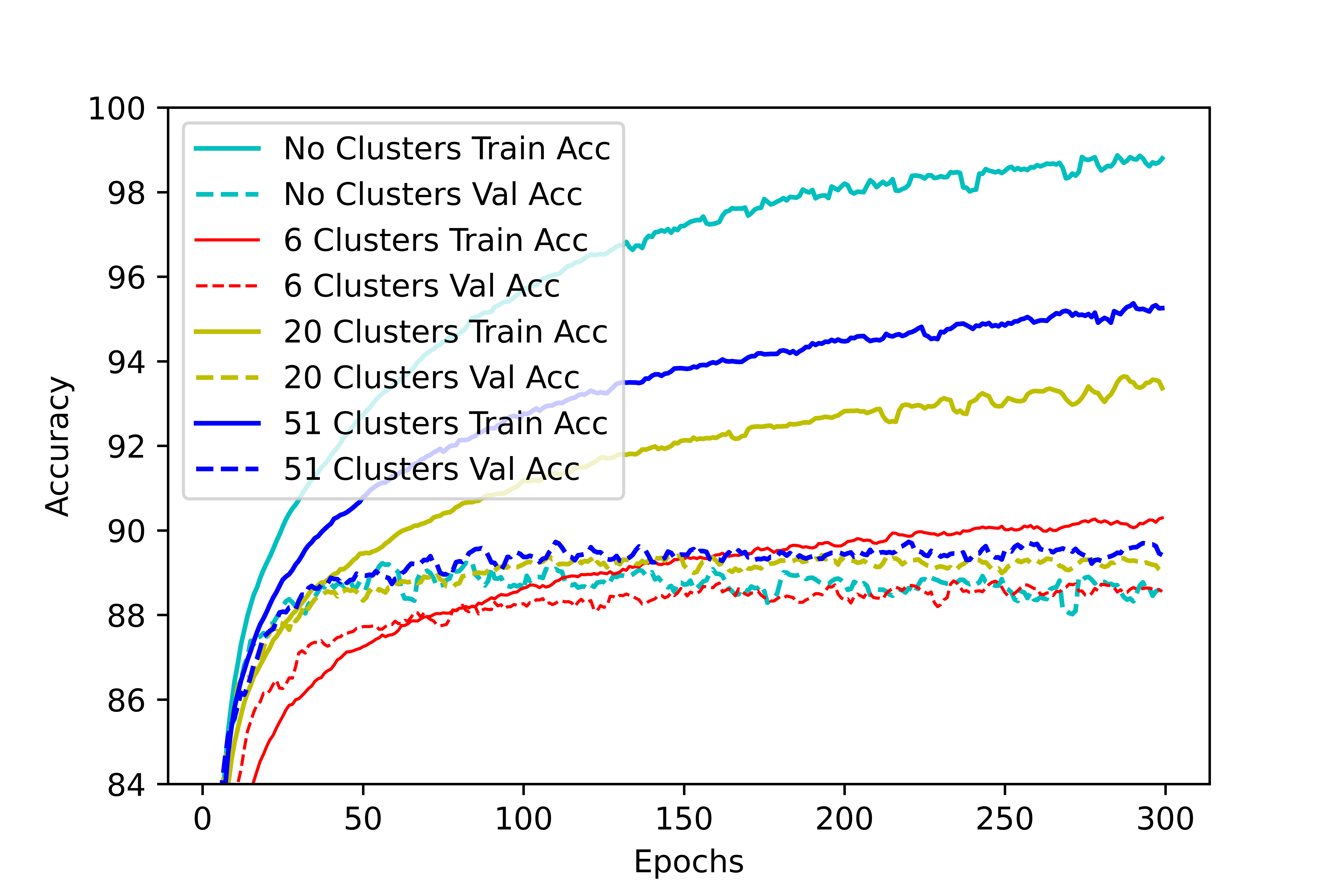}
    \includegraphics[width=0.49\linewidth]{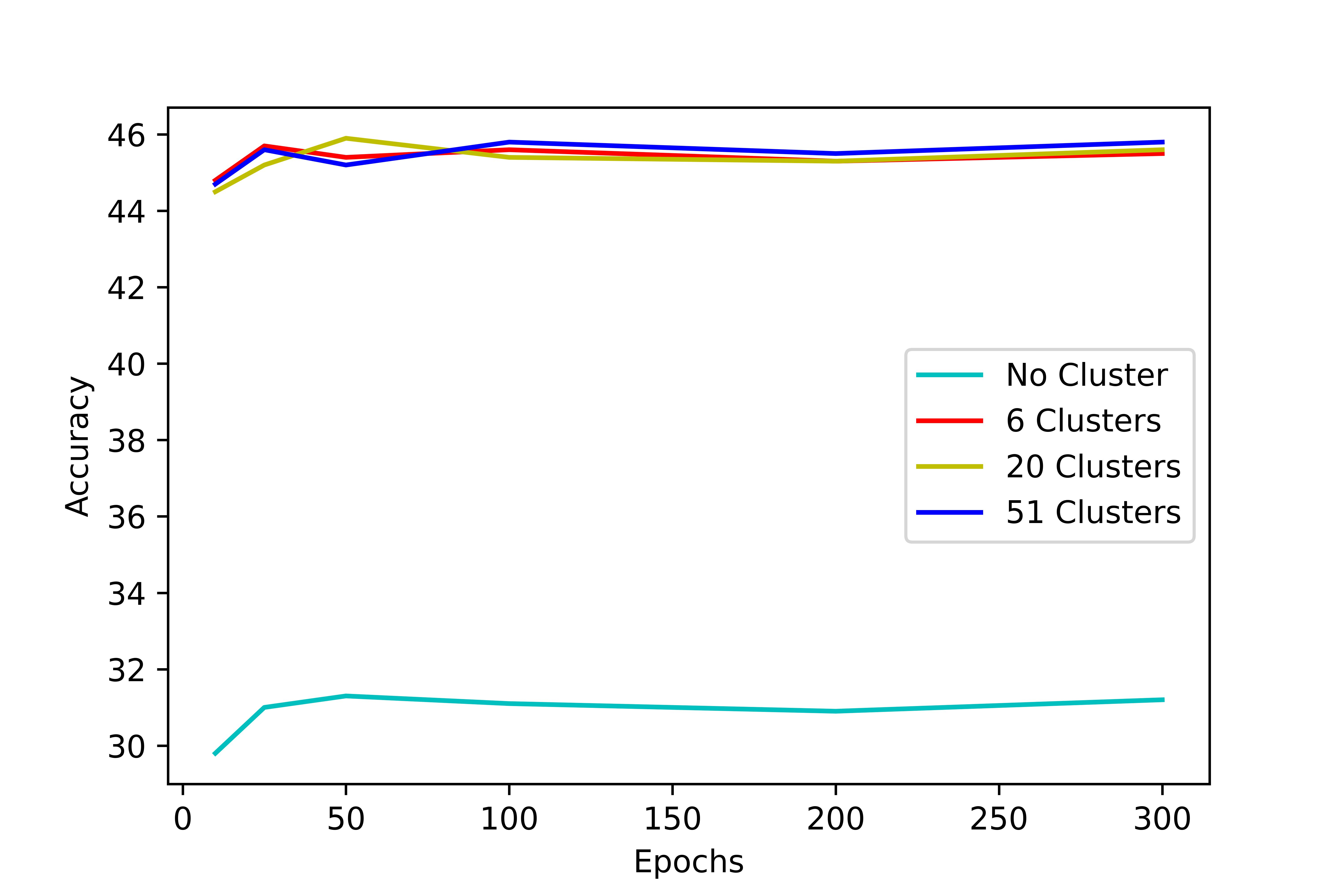}
    \caption{Left: Learning curve for the seen classes. Right: Learning curve for the unseen classes. The clustering-based representation avoids overfitting, which in the case of seen classes means that the gap between validation and training accuracy is smaller than in the vanilla representation. This regularization effect improves the validation accuracy in unseen classes. }
    \label{fig:seen_vs_unseen}
\end{figure}
Neural networks have proven extraordinarily powerful at learning complex discriminative functions of classes with many modes. In other words, instances of the same class can be very different and still be projected by the neural network to the same category. While this works well in supervised training, it can be a problem in Zero-Shot recognition, where the highly specialized discriminative function might not transfer well to instances of unseen classes. In this work, we adress this representation problem using three main ideas. 

First, we turn to clustering, and use the centroids of the clusters to  represent a video. We argue that centroids are more robust to outliers, and thus help regularize the representation, avoiding overfitting to the space of seen classes. Figure~\ref{fig:seen_vs_unseen} shows that the gap between training and validation accuracy is smaller when using clustering in seen classes (left). As a result, the learned representation is more general, which significantly improves accuracy in unseen classes (right). 
 
Second, our representation is a combination of a visual and a semantic representation. The standard practice at training time is to use a visual representation, and learn a mapping to the semantic representation. Instead, we use both cues, which we show yields a better representation. This is not surprising, since both visual and semantic information can complement each other. 


Third, we use the signal from classification as direct supervision for clustering, by using Reinforcement Learning (RL). Specifically, we use the REINFORCE algorithm to directly update the cluster centroids.
This optimization improves the clustering significantly 
and leads to less noisy and more compact representations for unseen classes.

These three pieces are essential to learn a robust, generalizable representation of videos for Zero-Shot action recognition, as we show in the ablation study. Crucially, none of them have been used in the context of Zero-Shot or action recognition. They are simple, yet fundamental and we hope they will be useful to anyone in the Zero-Shot and action recognition communities. 

We call the proposed method \MethodName, for \emph{CLustering with reinforcement learning for Action recognition in zero-ShoT lEaRning}, and show that it significantly outperforms all existing methods across all standard Zero-Shot action recognition datasets and tasks.

\section{Related Work}


\myparagraph{Fully Supervised Action Recognition.} 
This is the most widely studied setting in action recognition, where there is a large amount of samples at training time and the label spaces are the same at training and testing time. A thorough survey is beyond our scope, but as we make use of these in the backbone of our model, we mention some of the most widely used work. The seminal work of Simonyan and Zisserman \cite{simonyan2014two} introduced the now standard two-stream deep learning framework, which combines spatial and temporal information. Spatio-temporal CNNs~\cite{tran2015learning,qiu2017learning,i3d} are also widely used as backbones for many applications, including this work. More recently, research has incorporated attention~\cite{wang2018non,girdhar2017attentional} and leveraged the multi-modal nature of videos \cite{alwassel2019self}. In this work, we use the widely used I3D~\cite{i3d}. 
\\

\myparagraph{Zero-Shot Learning.}

Early approaches followed the idea of learning semantic classifiers for seen classes and then classifying the visual patterns by predicting semantic descriptions and comparing them with descriptions of unseen classes. In this space, Lampert et al.~\cite{lampert2009learning} propose attribute prediction, using the posterior of each semantic description. The SJE model~\cite{SJE} uses multiple compatibility functions to construct a joint embedding space. ESZSL~\cite{romera2015embarrassingly} uses a Frobenius norm
regularizer to learn an embedding space. Repurposing these methods for action classification is not trivial. In videos,
there are additional challenges: action labels need more complex representations than objects and hence give rise to more complex manual annotations. 
\\

\myparagraph{ZSL for Action Recognition.}
Early work \cite{rohrbach12eccv} was restricted to cooking activities, using script data to transfer to unseen classes.
Gan et al.~\cite{gan2016learning} consider each action class as a domain, and address semantic representation identification as a multi-source domain generalization problem.  
Manually specified semantic representations are simple and effective \cite{zhu2018towards} but labor-intensive to annotate. To overcome this, the use of label embeddings has proven popular, as only category names are needed. Some approaches use common embedding space between class labels and video features~\cite{xu2016multi,xu2017transductive}, pairwise relationships between classes \cite{gan2016concepts}, error-correcting codes~\cite{qin2017zero}, inter-class relationships \cite{gan2015exploring}, out-of-distribution detectors~\cite{OD}, and Graph Neural networks~\cite{psgnn}. In contrast, we are learning to optimize centroids of visual semantic representations that generalize better to unseen classes. 
\\ 

\begin{figure*}[t]
    \centering
    \includegraphics[width=0.9\linewidth]{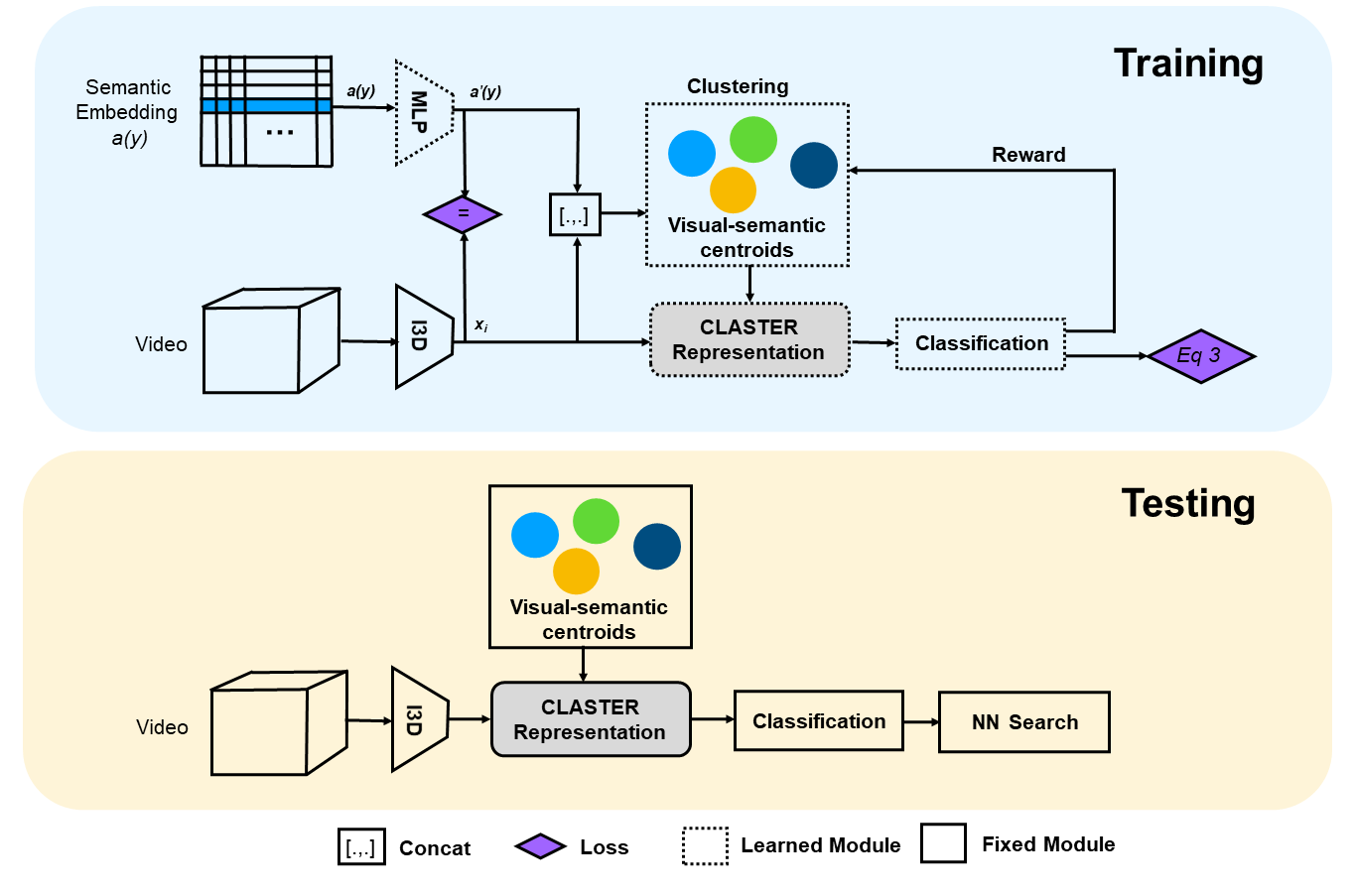}
    \caption{Overview of \MethodName. We map the semantic embedding $a(y_i)$ to the space of visual features $x_i$, and concatenate both to obtain a visual-semantic representation. We cluster these visual-semantic representations with K-means to obtain initial cluster centroids. Each video is represented as the sum of the visual-semantic representation and the centroid clusters, weighted by their distance (see Sec.~\ref{claster_rep} and Fig.~\ref{fig:c2f}). 
    This is used as input for classification (Sec~\ref{sec:loss} and Eq. \ref{loss}). Based on the classification result, we send a reward to optimize the cluster centroids using REINFORCE (Sec.~\ref{subsec:RL}). At test time, we first perform classification on the seen classes and then do a nearest neighbor (NN) search to predict the unseen class.}
    \label{fig:overview}
\end{figure*}

\myparagraph{Reinforcement Learning for Zero-Shot Learning.}

RL for ZSL in images was introduced by Liu et al.~\cite{liu2018combining} by using a combination of ontology and RL. 
In Zero-Shot text classification, Ye et al.~\cite{ye2020zero} propose a self-training method
to leverage unlabeled data. 
RL has also been used in the Zero-Shot setting for task generalization \cite{oh2017zero}, active learning \cite{fan2017learning}, and video object segmentation \cite{gowda2020alba}.
To the best of our knowledge, there is no previous work using RL for optimizing centroids in Zero-Shot  recognition. 
\\

\myparagraph{Deep Approaches to Centroid Learning for Classification.}

Since our approach learns cluster centroids using RL, it is related to the popular cluster learning strategy for classification called Vector of Locally Aggregated Descriptors (VLAD) \cite{netvlad}. The more recent NetVLAD~\cite{netvlad} leverages neural networks which helps outperform the  standard VLAD by a wide margin. ActionVLAD~\cite{actionvlad} aggregates NetVLAD over time to obtain descriptors for videos. ActionVLAD uses clusters that correspond to spatial locations in a video while we use joint visual semantic embeddings for the entire video. In general, VLAD uses residuals
with respect to cluster centroids as representation while \MethodName uses a weighting of the centroids. 
The proposed \MethodName outperforms NetVLAD by a large margin on both HMDB51 and UCF101.

\section{\MethodName}

We now describe the proposed \MethodName, which leverages clustering of visual and semantic features for video action recognition and optimizes the clustering with RL. Figure~\ref{fig:overview} shows an overview of the method. 

\subsection{Problem Definition}


Let $S$ be the training set of seen classes. $S$ is composed of tuples $\left( x, y, a(y) \right)$, where $x$ 
represents the spatio-temporal features of a video,
$y$ represents the class label in the set of $Y_{S}$ seen class labels, and $a(y)$ denotes the category-specific semantic representation of class $y$. These semantic representations are either manually annotated 
or
computed using a language-based embedding of the category name, such as word2vec~\cite{word2vec} or sentence2vec~\cite{sentence2vec}. 


Let $U$ be the set of pairs $ \left (u, a(u)\right)$, where $u$ is a class in the set of unseen classes $Y_{U}$ and $a(u)$ are the corresponding semantic representations. The seen classes $Y_S$ and the unseen classes $Y_U$ do not overlap. 

In the Zero-Shot Learning (ZSL) setting, given an input video the task is to predict a class label in the unseen classes, as $f_{ZSL}:X\rightarrow Y_{U}$. In the related generalized Zero-Shot learning (GZSL) setting, given an input video, the task is to predict a class label in the union of the seen and unseen classes, as $f_{GZSL}:X\rightarrow Y_{S}\cup Y_{U}$. 

\subsection{Visual-Semantic Representation}

Given video $i$, we compute visual features $x_{i}$ and a semantic embedding $a(y_i)$ of their class $y_i$ (see Sec.~\ref{subsec:implementation} for details). The goal is to map both to the same space, so that they have the same dimensionality and magnitude, and therefore they will have a similar weight during clustering. We learn this mapping with a simple multi-layer perceptron (MLP)~\cite{zhang2017learning}, 
trained with a least-square loss. This loss minimizes the distance between $x_{i}$ and the output from the MLP, which we call $a'(y)$. 
Finally, we concatenate $x_{i}$ and $a'(y)$ to obtain the visual-semantic representations that will be clustered. The result is a representation which is not only aware of the visual information but also the semantic. 



\subsection{CLASTER Representation}
\label{claster_rep}
We now detail how we represent videos using the 
proposed \emph{CLASTER} representation, which leverages clustering as a form of regularization. 
In other words, a representation w.r.t~centroids is more robust to outliers, which is helpful since all instances of the unseen classes are effectively outliers w.r.t. the training distribution. 

\begin{figure}[t]
    \centering
    \includegraphics[width=0.7\linewidth]{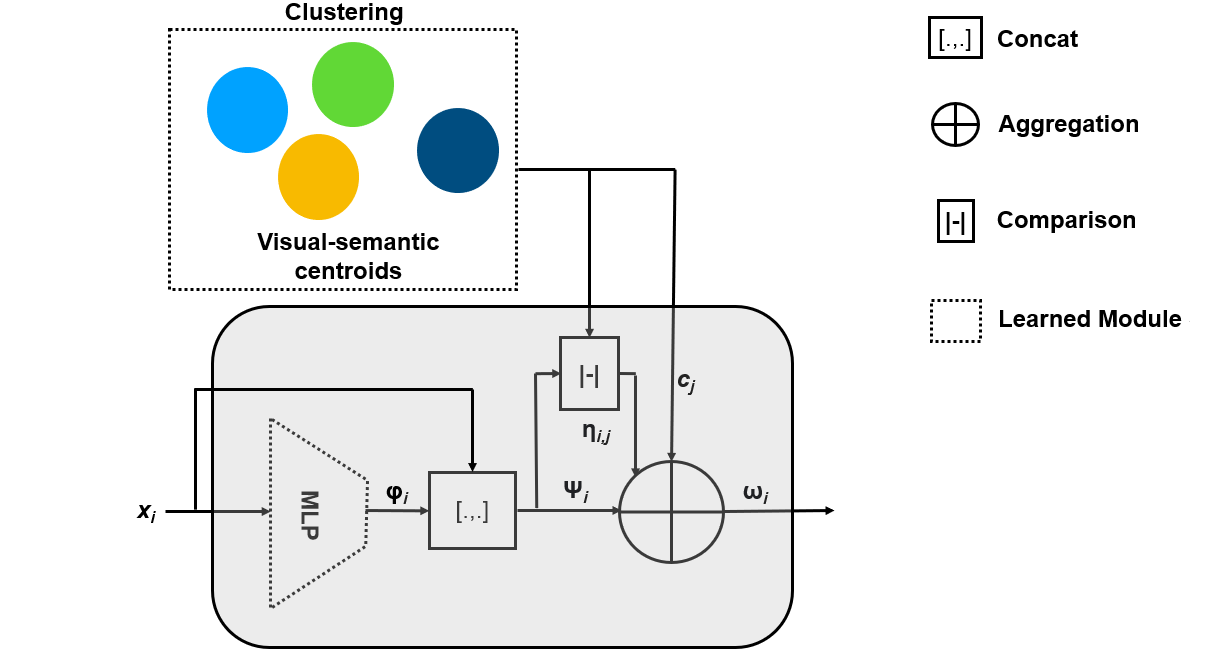}
    \vspace{-3pt}
    \caption{The proposed CLASTER Representation in detail (see Fig.~\ref{fig:overview} for the overview of the full method). The visual feature is mapped to match the space of the visual-semantic cluster centroids with an MLP and concatenation. Based on the distances to the cluster centroids the final representation $\omega$ is a weighted representation of the centroids, more robust to the out-of-distribution instances of the unseen test classes. Details in Sec.~\ref{claster_rep} and Eq.~\ref{eq:inter}.}
    \label{fig:c2f}
\end{figure}

We initialize the clustering of 
the training set $S$ using K-means~\cite{forgy1965cluster}. Each resulting cluster $j$ 
has a centroid $c_{j}$, that is the average of all visual-semantic samples in that particular cluster. The \MethodName representation of a given video is the sum 
of the visual-semantic representation and 
the centroids, weighted by the inverse of the distance, such that closer clusters will have more weight. Figure~\ref{fig:c2f} shows this process in detail. 

Specifically, given video $i$, we compute the visual representation $x_{i}$. We estimate the semantic vector $\phi_{i}$ using an MLP. This is a necessary step, as during test time we do not have any semantic information. Concatenating the visual $x_{i}$ and semantic $\phi_{i}$ we obtain the intermediate representation $\psi_{i}$, which is in the same space as the cluster centroids. 

We compute the Euclidean distance $d_{i,j}$ between the visual-semantic point $\psi_{i}$ and each cluster $j$, which we refer to as $d_{i,j}$. We take the inverse $1/d_{i,j}$ and normalize them using their maximum and minimum values, such that they are between 0 and 1. We refer to these normalized values as $\eta_{i,j}$, and they are used as the weights of each cluster centroid in the final \MethodName representation $\omega_{i}$:
\begin{equation}
    \omega_{i}= \psi_{i} + \sum_{j=1}^{k}\eta_{i,j}c_{j}.
    \label{eq:inter}
\end{equation}



\subsection{Loss Function}
\label{sec:loss}

Given the \MethodName representation $\omega_i$ we predict a seen class using a simple MLP,  $V$. 
Instead of the vanilla softmax function, we use semantic softmax~\cite{ji2018semantic}, 
which includes the semantic information $a(y_i)$ and thus can transfer better to Zero-Shot classes:
\begin{equation}
    \label{softmax}
   \hat y_{i} =\frac{e^{a(y_{i})^{T}V(\omega_{i})}}{\sum_{j=1}^{S}e^{a(y_j)^{T}V(\omega_{i})}}.
\end{equation}

The output $\hat y_{i}$ is a vector with a probability distribution over the $S$ seen classes. 
We train the classifier, which minimizes the cross-entropy loss with a regularization term: 

\begin{equation}
    \label{loss}
    \textup{min}_{W}\sum_{i=1}^{N}\mathcal{L}(x_{i})+\lambda \left \| W \right \|,
\end{equation}

where $W$ refers to all weights in the network.



\subsection{Optimization with Reinforcement Learning}
\label{subsec:RL}


As there is no ground truth for clustering centroids, we use the classification accuracy as supervision. This makes the problem non-differentiable, and therefore we cannot use traditional gradient descent. Instead, we use RL to optimize the cluster centroids.
For this, we compute two variables that will determine each centroid update: 
the reward, which measures whether the classification is correct and will determine the direction of the update, and the classification score, which measures how far the prediction is from the correct answer and will determine the magnitude of the update. 

Given the probabilistic prediction $\hat y_{i}$ and the one-hot representation of the ground truth class $y_{i}$, we compute the classification score as the dot product of the two: $z_{i}=y_{i} \hat y_{i}$. To obtain the reward, we check if the maximum of $\hat y_{i}$ and $y_{i}$ lie in the same index: 

\begin{equation}
\label{reward}
    r = \left\{\begin{matrix}
1 \:\: if \:\: \arg\max \hat y_{i} =  \arg\max y_{i} \\ 
-1 \:\: otherwise \hspace{2.8cm}\\ 
\end{matrix}\right.
\end{equation} 

This essentially gives a positive reward if the model has predicted a correct classification and a negative reward if the classification was incorrect. This formulation is inspired by Likas \cite{likas1999reinforcement}, which was originally proposed for a different domain and the problem of competitive learning. 

For each data point $\psi_{i}$ we only update the closest cluster centroid $c_{j}$. We compute the update $\Delta c_{j}$ using the REINFORCE \cite{likas1999reinforcement} algorithm as:









\begin{equation}
    \label{update}
    \Delta c_{j} = \alpha\  r\ (z_{i}-p_{j})\ (\psi_{i}-c_{j}).
\end{equation}

For further details on this derivation, please see the Supplementary Material as well as Likas~\cite{likas1999reinforcement}. The main  difference between our model and Likas' is that we do not consider cluster updates to be Bernoulli units. 
Instead, we modify the cluster centroid with the classification score $z_{i}$, which is continuous in the range between 0 and 1.

\section{Implementation Details} 
\label{subsec:implementation}

\myparagraph{Visual features.} We use RGB and flow features extracted from the \textit{Mixed 5c} layer of an I3D network pre-trained on the Kinetics~\cite{i3d} dataset. The \textit{Mixed 5c} output of the flow network is averaged across the temporal dimension and pooled by four in the spatial dimension and then flattened to a vector of size 4096. We then concatenate the two. 
\\

\myparagraph{Network architecture.} The MLP that maps the semantic features to visual features consists of two fully-connected (FC) layers and a ReLU. The MLP in the CLASTER Representation module, which maps the visual feature to the semantic space is a two-layer FC network, whose output after concatenation with the video feature has the same dimensions as the cluster representatives. The size of the FC layers is 8192 each. The final classification MLP (represented as a classification block in Figure~\ref{fig:overview}) consists of two convolutional layers and two FC layers, where the last layer equals the number of seen classes in the dataset we are looking at. All the modules are trained with the Adam optimizer with a learning rate of 0.0001 and weight
decay of 0.0005.
\\ 

\myparagraph{Number of clusters.} Since the number of clusters is a hyperparameter, we evaluate the effect of the number of clusters on the UCF101 dataset for videos and choose 6 after the average performance stablizes as can be seen in the supplementary material. We then use the same number for the HMDB51 and Olympics datasets. 
\\ 

\myparagraph{RL optimization.} We use 10,000 iterations and the learning rate $\alpha$ is fixed to 0.1 for the first 1000 iterations, 0.01 for the next 1000 iterations and then drop it to 0.001 for the remaining iterations.
\\ 


\myparagraph{Semantic embeddings.} We experiment with three types of embeddings as semantic representations of the classes. We have human-annotated semantic representations for UCF101 and the Olympic sports dataset of sizes 40 and 115 respectively. HMDB51 does not have such annotations. 
Instead, we use a skip-gram model trained on the news corpus provided by Google to generate word2vec embeddings. Using action classes as input, we obtain a vector representation of 300 dimensions.
Some class labels contain multiple words. In those cases, we use the average of the word2vec embeddings. We also use sentence2vec embeddings, trained on Wikipedia. These can be obtained for both single words and multi-word expressions. The elaborate descriptions are taken from \cite{er} and only evaluated for fair comparison to them.\\
For the elaborative descriptions, we follow ER~\cite{er} and use the provided embeddings in their codebase.
\\



\myparagraph{Rectification of the Semantic Embedding}

Sometimes, in ZSL, certain data points tend to appear as nearest-neighbor of many other points in the projection space. This is referred to as the hubness problem \cite{shigeto2015ridge}. We avoid this problem using semantic rectification~\cite{luo2017zero}, where the class representation is modified by averaging the output generated by the projection network, which in our case is the penultimate layer of the classification MLP. 
Specifically, for the unseen classes, we perform rectification by first using the MLP 
trained on the seen classes to project the semantic embedding to the visual space. 
We add the average of projected semantic embeddings from the k-nearest neighbors of the seen classes, specifically as follows:

\begin{equation}
    \label{attr}
    \hat a(y_{i})=a'(y_{i})+\frac{1}{k}\sum_{n\in N}cos \left( a'(y_{i}),n \right ) \cdot n,
\end{equation}


where $a'(y)$ refers to the embedding after projection to the visual space, $cos(a,n)$ refers to the cosine similarity between $a$ and $n$, the operator $\cdot$ refers to the dot product and $N$ refers to the k-nearest neighbors of $a'(y_{u_{i}})$.
\\

\myparagraph{Nearest Neighbor Search}
At test time in ZSL, given a test video, we predict a seen class and compute or retrieve its semantic representation. After rectification, we find the nearest neighbor in the set of unseen classes. 
%
%
In the GZSL task, class predictions may be of seen or unseen classes. Thus, we first use a bias detector~\cite{gao2019know} which helps us detect if the video belongs to the seen or unseen class. If it belongs to a seen class, we predict the class directly from our model, else we proceed as in ZSL. 
\\

\section{Experimental Analysis}
\label{sec:analysis}

In this section, we look at the qualitative and quantitative performance of the proposed model. We first describe the experimental settings, and then show an ablation study, that explores the contribution of each component. We then compare the proposed method to the state-of-the-art in the ZSL and GZSL tasks, and give analytical insights into the advantages of \MethodName. 


\subsection{Datasets}

We choose the Olympic Sports \cite{olympics}, HMDB-51 \cite{hmdb} and UCF-101 \cite{ucf101}, so that we can compare to recent state-of-the-art models \cite{gan2016learning,OD,qin2017zero}. 
We follow the commonly used 50/50 splits of Xu et al. \cite{xu2017transductive}, where 50 percent are seen classes and 50 are unseen classes. Similar to previous approaches \cite{zhu2018towards,gan2016learning,qin2017zero,mettes2017spatial,kodirov2015unsupervised}, we report average accuracy and standard deviation over 10 independent runs. 
We report results on the split proposed by \cite{xian2017zero}, in the standard inductive setting. We also report on the recently introduced TruZe~\cite{truze}. This split accounts for the fact that some classes present on the dataset used for pre-training (Kinetics~\cite{i3d}) overlap with some of the unseen classes in the datasets used in the Zero-Shot setting, therefore breaking the premise that those classes have not been seen. 

\subsection{Ablation Study}
\label{subsec:ablation}



Table~\ref{tab:ablation} shows the impact of using the different components of \MethodName. \\

\myparagraph{Impact of Clustering.} We consider several baselines. First, omitting clustering, which is the equivalent of setting $\omega_{i}= \psi_{i}$, in Eq.~\ref{eq:inter}. This is, ignoring the cluster centroids in the representation. This is referred to in Table~\ref{tab:ablation} as ``No clustering".  Second, we use random clustering, which is assigning each instance to a random cluster. Finally, we use the standard K-means. We observe that using clusters is beneficial, but only if they are meaningful, as in the case of K-means. \\
    
\myparagraph{Impact of using a visual-semantic representation.} We compare to the standard representation, which only includes visual information, and keep everything the same. This is, clustering and classification are done using only the visual features. This is referred to in the table as ``\MethodName w/o SE". We observe that there is a very wide gap between using and not using the semantic features at training time. This effect is present across all datasets, suggesting it is a general improvement in the feature learning. We also show a comparison of aggregation strategies and interaction between visual and semantic features in the supplementary material.\\
    
\myparagraph{Impact of different optimization choices.} We make cluster centroids learnable parameters and use the standard SGD to optimize them (`\MethodName w/o RL") 
We also test the use of the related work of NetVLAD to optimize the cluster (``\MethodName w/ NetVLAD"). 
We see that the proposed model outperforms NetVLAD by an average of 4.7\% and the CLASTER w/o RL by 7.3\% on the UCF101 dataset. A possible reason for this difference is that the loss is back-propagated through multiple parts of the model before reaching the centroids. However, with RL the centroids are directly updated using the reward signal. 
Section~\ref{sec:analysis:RL} explores how the clusters change after the RL optimization. In a nutshell, the RL optimization essentially makes the clusters cleaner, moving most instances in a class to the same cluster.



\setlength{\tabcolsep}{1.5pt}
\begin{table}
\begin{center}
\begin{tabular}{|l|c|c|c|}
\hline
Component  & HMDB51 & Olympics & UCF101\\
\hline\hline
No clustering & 25.6 $\pm$ 2.8 &  57.7 $\pm$ 3.1 & 31.6 $\pm$ 4.6\\
Random clustering (K=6)& 20.2 $\pm$ 4.2 & 55.4 $\pm$ 3.1& 24.1 $\pm$ 6.3 \\
K-means (K=6)  & 27.9 $\pm$ 3.7 & 58.6 $\pm$ 3.5& 35.3 $\pm$ 3.9\\
\hline


\MethodName w/o SE & 27.5 $\pm$ 3.8 & 55.9 $\pm$ 2.9 & 39.4 $\pm$ 4.4\\
\hline
\MethodName w/o RL  & 30.1  $\pm$ 3.4 & 60.5  $\pm$ 1.9& 39.1 $\pm$ 3.2\\
\MethodName w/ NetVLAD  & 33.2  $\pm$ 2.8 & 62.6  $\pm$ 4.1& 41.7 $\pm$ 3.8\\
\hline
\MethodName  & {\bf 36.8 $\pm$ 4.2} & {\bf 63.5 $\pm$ 4.4}& {\bf 46.4 $\pm$ 5.1}\\
\hline
\end{tabular}
\end{center}
\caption{Results of the ablation study of different components of \MethodName ZSL. 
The study shows the effect of clustering, using visual-semantic representations, and optimizing with different methods. All three components show a wide improvement over the various baselines, suggesting that they are indeed complementary to improve the final representation. 
} 
\label{tab:ablation}
\end{table}

\subsection{Results on ZSL}
\label{ZSL}

Table~\ref{tbl:results:zsl} shows the comparison between \MethodName and several state-of-the-art methods: the out-of-distribution detector method (OD)~\cite{OD}, a generative approach to Zero-Shot action recognition (GGM)~\cite{GGM2018}, the evaluation of output embeddings (SJE)~\cite{SJE}, the feature generating networks (WGAN)~\cite{clswgan}, the end-to-end training for realistic applications approach (E2E)~\cite{e2e}, the inverse autoregressive flow (IAF) based generative model, bi-directional adversarial GAN(Bi-dir GAN) \cite{syn} and prototype sampling graph neural network (PS-GNN)~\cite{psgnn}. To make results directly comparable, we use the same backbone across all of them, which is the I3D~\cite{i3d} pre-trained on Kinetics.

We observe that the proposed \MethodName consistently outperforms all other state-of-the-art methods across all datasets. The improvements are significant: up to 3.5\% on HMDB51 and 13.5\% on UCF101 with manual semantic embedding. 
We also measure the impact of different semantic embeddings, including using sentence2vec instead of word2vec. 
We show that sentence2vec significantly improves over using word2vec, especially on UCF101 and HMDB51. Combination of embeddings resulted in average improvements of 0.3\%, 0.8\% and 0.9\% over the individual best performing embedding of \MethodName.
 
 \newcommand{\cent}[1]{\multicolumn{1}{c|}{#1}}
\setlength{\tabcolsep}{2pt}
\begin{table}[htb]
\small
\begin{center}
\begin{tabular}{|l|c|c|c|c|}
\hline
Method & SE & Olympics & HMDB51 & UCF101\\
\hline\hline
SJE \cite{SJE}& M &  47.5 $\pm$ 14.8 & \cent{-} & 12.0 $\pm$ 1.2 \\
Bi-Dir GAN \cite{syn} & M & 53.2 $\pm$ 10.5 & \cent{-} &	24.7 $\pm$ 3.7\\
IAF \cite{syn} & M & 54.9 $\pm$ 11.7 & \cent{-} & 26.1 $\pm$ 2.9 \\

GGM \cite{GGM2018}& M & 57.9 $\pm$ 14.1 & \cent{-} & 24.5 $\pm$ 2.9 \\
OD \cite{OD} & M & 65.9 $\pm$ 8.1 & \cent{-} & 38.3 $\pm$ 3.0\\
WGAN \cite{clswgan} & M & 64.7 $\pm$ 7.5 & \cent{-} & 37.5 $\pm$ 3.1 \\

\textbf{\MethodName (ours)} & M & \textbf{67.4 $\pm$ 7.8} & \cent{-} & \textbf{51.8 $\pm$ 2.8} \\
\hline
SJE \cite{SJE}& W &  28.6 $\pm$ 4.9 & 13.3 $\pm$ 2.4 & 9.9 $\pm$ 1.4 \\
IAF \cite{syn} & W & 39.8 $\pm$ 11.6 & 19.2 $\pm$ 3.7 & 22.2 $\pm$ 2.7 \\
Bi-Dir GAN \cite{syn} & W & 40.2 $\pm$ 10.6 & 21.3 $\pm$ 3.2 & 21.8 $\pm$ 3.6\\
GGM \cite{GGM2018} & W & 41.3 $\pm$ 11.4 & 20.7 $\pm$ 3.1 & 20.3 $\pm$ 1.9 \\

WGAN \cite{clswgan} & W & 47.1 $\pm$ 6.4 & 29.1 $\pm$ 3.8 & 25.8 $\pm$ 3.2 \\
OD \cite{OD} & W & 50.5 $\pm$ 6.9 & 30.2 $\pm$ 2.7 & 26.9 $\pm$ 2.8 \\
PS-GNN \cite{psgnn} & W & 61.8 $\pm$ 6.8 & 32.6 $\pm$ 2.9 & 43.0 $\pm$ 4.9 \\
E2E \cite{e2e}* & W & 61.4 $\pm$ 5.5 & 33.1 $\pm$ 3.4  & 46.2 $\pm$ 3.8 \\
\textbf{\MethodName (ours)}  & W & \textbf{63.8 $\pm$ 5.7} & \textbf{36.6 $\pm$ 4.6} & \textbf{46.7 $\pm$ 5.4} \\

\hline

\textbf{\MethodName (ours)}  & S & \textbf{64.2 $\pm$ 3.3}  & \textbf{41.8 $\pm$ 2.1} & \textbf{50.2 $\pm$ 3.8} \\
\hline
\textbf{\MethodName (ours)} & C & \textbf{67.7 $\pm$ 2.7}  & \textbf{42.6 $\pm$ 2.6} & \textbf{52.7 $\pm$ 2.2} \\
\hline
ER \cite{er} & ED & 60.2 $\pm$ 8.9 & 35.3 $\pm$ 4.6 & 51.8 $\pm$ 2.9 \\
\textbf{\MethodName (ours)} & ED & \textbf{ 68.4 $\pm$ 4.1 }  & \textbf{ 43.2 $\pm$ 1.9 } & \textbf{53.9 $\pm$ 2.5 } \\
\hline
\end{tabular}
\end{center}
\caption{Results on ZSL. SE: semantic embedding, M: manual representation, W: word2vec embedding, S: sentence2vec, C: Combination of embeddings. The proposed \MethodName outperforms previous state-of-the-art across tasks and datasets. 
}
\label{tbl:results:zsl}
\end{table}


\subsection{Results on GZSL}
\label{GZSL}

We now compare to the same approaches in the GZSL task in Table~\ref{tbl:results:gzsl},  the reported results are the harmonic mean of the seen and unseen class accuracies. Here \MethodName outperforms all previous methods across different modalities. We obtain an improvement on average of 2.6\% and 5\% over the next best performing method on the Olympics dataset using manual representations and word2vec respectively. We obtain an average improvement of 6.3\% over the next best performing model on the HMDB51 dataset using word2vec. We obtain an improvement on average performance by 1.5\% and 4.8\% over the next best performing model on the UCF101 dataset using manual representations and word2vec respectively. Similarly to ZSL, we show generalized performance improvements using sentence2vec. We also report results on the combination of embeddings. We see an improvement of 0.3\%, 0.6\% and 0.4\% over the individual best  embedding for \MethodName. The seen and unseen accuracies are shown in the Supplemental Material.

\setlength{\tabcolsep}{2pt}
\begin{table}
\small
\begin{center}
\begin{tabular}{|l|c|c|c|c|}
\hline
Method & SE & Olympics & HMDB51 & UCF101\\
\hline\hline
Bi-Dir GAN \cite{syn} & M & 44.2 $\pm$ 11.2 & \cent{-} &	22.7 $\pm$ 2.5\\
IAF \cite{syn} & M & 48.4 $\pm$ 7.0 & \cent{-} &	25.9 $\pm$ 2.6\\

GGM \cite{GGM2018} & M & 52.4 $\pm$ 12.2  & \cent{-} & 23.7 $\pm$ 1.2 \\
WGAN \cite{clswgan} & M & 59.9 $\pm$ 5.3 & \cent{-} & 44.4 $\pm$ 3.0 \\
OD\cite{OD}  & M & 66.2 $\pm$ 6.3 & \cent{-} & 49.4 $\pm$ 2.4\\

\textbf{\MethodName (ours)}  & M & \textbf{68.8 $\pm$ 6.6 } & \cent{-} & \textbf{50.9 $\pm$ 3.2} \\
\hline
IAF \cite{syn} & W & 30.2 $\pm$ 11.1	& 15.6 $\pm$ 2.2 & 20.2 $\pm$ 2.6	\\
Bi-Dir GAN \cite{syn} & W & 32.2 $\pm$ 10.5 & \ \ 7.5 $\pm$ 2.4 & 17.2 $\pm$ 2.3\\
SJE  \cite{SJE}& W & 32.5 $\pm$ 6.7 & 10.5 $\pm$ 2.4 & \ \ 8.9 $\pm$ 2.2 \\
GGM\cite{GGM2018}  & W & 42.2 $\pm$ 10.2 & 20.1 $\pm$ 2.1 & 17.5 $\pm$ 2.2 \\

WGAN \cite{clswgan} & W & 46.1 $\pm$ 3.7 & 32.7 $\pm$ 3.4 & 32.4 $\pm$ 3.3 \\
PS-GNN \cite{psgnn} & W & 52.9 $\pm$ 6.2 & 24.2 $\pm$ 3.3 & 35.1 $\pm$ 4.6 \\
OD \cite{OD} & W & 53.1 $\pm$ 3.6  & 36.1 $\pm$ 2.2 & 37.3 $\pm$ 2.1 \\

\textbf{\MethodName (ours)}  & W & \textbf{58.1 $\pm$ 2.4 } & \textbf{42.4 $\pm$ 3.6} & \textbf{42.1 $\pm$ 2.6} \\

\hline

\textbf{\MethodName (ours)}  & S & \textbf{58.7 $\pm$ 3.1} & \textbf{47.4 $\pm$ 2.8} & \textbf{48.3 $\pm$ 3.1} \\
\hline
\textbf{\MethodName (ours)}  & C & \textbf{69.1 $\pm$ 5.4} & \textbf{48.0 $\pm$ 2.4} & \textbf{51.3 $\pm$ 3.5} \\
\hline
\end{tabular}
\end{center}
\caption{Results on GZSL. SE: semantic embedding, M: manual representation, W: word2vec embedding, S: sentence2vec, C: combination of embeddings. The seen and unseen class accuracies are listed in the supplementary material. 
}
\label{tbl:results:gzsl}
\end{table}

\subsection{Results on TruZe}


We also evaluate on the more challenging TruZe split. The proposed UCF101 and HMDB51 splits have 70/31 and 29/22 classes (represented as training/testing). 
We compare to WGAN \cite{clswgan}, OD \cite{OD} and E2E \cite{e2e} on both ZSL and GZSL scenarios. Results are shown in Table~\ref{tbl:truze}.


\begin{table}[htb]
\small
\begin{center}
\begin{tabular}{| *{5}{c|} }
\hline
Method & \multicolumn{2}{c|}{UCF101 } & \multicolumn{2}{c|}{HMDB51}\\
 & ZSL & GZSL & ZSL & GZSL \\
 \hline\hline
WGAN & 22.5 & 36.3 & 21.1 & 31.8 \\
OD & 22.9 & 42.4 & 21.7 & 35.5 \\
E2E & 45.5 & 45.9 & 31.5 & 38.9 \\
\textbf{\MethodName} & \textbf{45.8} & \textbf{47.3} & \textbf{33.2} & \textbf{44.5} \\
\hline
\end{tabular}
\end{center}
\caption{Results on TruZe. For ZSL, we report the mean class accuracy and for GZSL, we report the harmonic mean of seen and unseen class accuracies. All approaches use sen2vec annotations as the form of semantic embedding. }
\label{tbl:truze}
\end{table}


\subsection{Analysis of the RL optimization}
\label{qualitative}
\label{sec:analysis:RL}

\begin{figure}[t]
    \centering
    \includegraphics[width=0.7\linewidth]{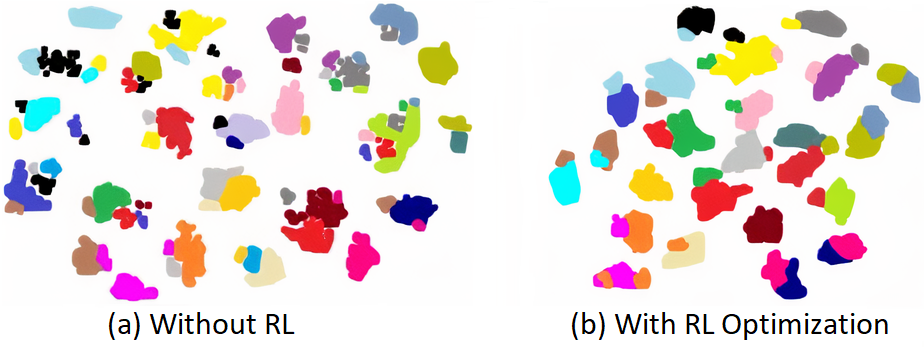}
    \caption{\MethodName improves the representation and clustering in unseen classes. The figure shows t-SNE  \cite{tsne} of video instances, where each color corresponds to a unique unseen class label. The RL optimization improves the representation by making it more compact: in (b) instances of the same class, i.e. same color, are together and there are less outliers for each class compared to (a).}
    \label{fig:teaser}
\end{figure}

We analyze how optimizing with RL affects clustering on the UCF101 training set. Figure~\ref{fig:teaser} shows the t-SNE~\cite{tsne} visualization. Each point is a video instance in the unseen classes, and each color is a class label. As it can be seen, the RL optimization makes videos of the same class appear closer together.

We also do a quantitative analysis of the clustering. For each class in the training set, we measure the distribution of clusters that they belong to, visualized in the Fig~\ref{fig:histogram}. We observe that after the RL optimization, the clustering becomes ``cleaner". This is, most instances in a class belong to a dominant cluster. This effect can be measured using the purity of the cluster: 

\begin{figure}
    \centering
    \includegraphics[width=0.87\linewidth]{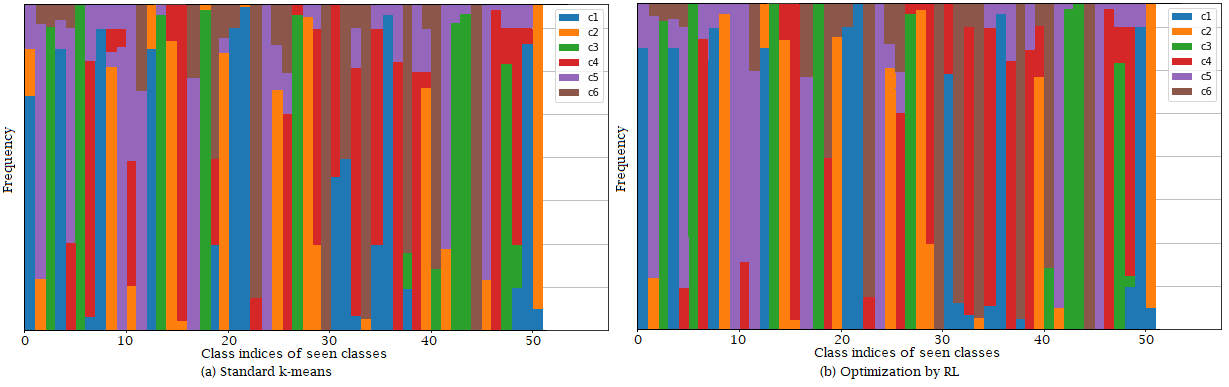}
    \caption{Analysis of how RL optimization changes the cluster to which an instance belongs. The frequencies are represented as percentages of instances in each cluster. We can see that the clusters are a lot "cleaner" after the optimization by RL.}
    \label{fig:histogram}
\end{figure}

\begin{equation}
    \label{purity}
    Purity = \frac{1}{N}\sum_{i=1}^{k}max_{j}\left |c_{i} \cap t_{j} \right |, 
\end{equation}

where $N$ is the number of data points (video instances), $k$ is the number of clusters, $c_{i}$ is a cluster in the set of clusters, and $t_{j}$ is the class which has the maximum count for cluster $c_{i}$.  
Poor clustering results in purity values close to 0, and a perfect clustering will return a purity of 1. Using K-means, the purity is 0.77, while optimizing the clusters with RL results in a purity of 0.89.

Finally, we observe another interesting side effect of clustering. Some of the most commonly confused classes before clustering (e.g.~``Baby crawling" vs.~``Mopping floor", ``Breaststroke" vs.~``front crawl", ``Rowing vs. front crawl") are assigned to different clusters after RL, resolving confusion. This suggests that clusters are also used as a means to differentiate between similar classes. 

\section{Regularization Effect of Clustering}
\label{sec:graphs}
In the main paper, we showed the regularization effect that clustering had when using 6, 10 and 51 clusters in comparison to no clusters. Here, we look at the same effect with 20 and 35 clusters as well. We see consistent improvements of over 15\% in accuracy for the unseen classes using the proposed CLASTER representation compared to no clustering.
\begin{figure}
    \centering
    \includegraphics[width=0.49\linewidth]{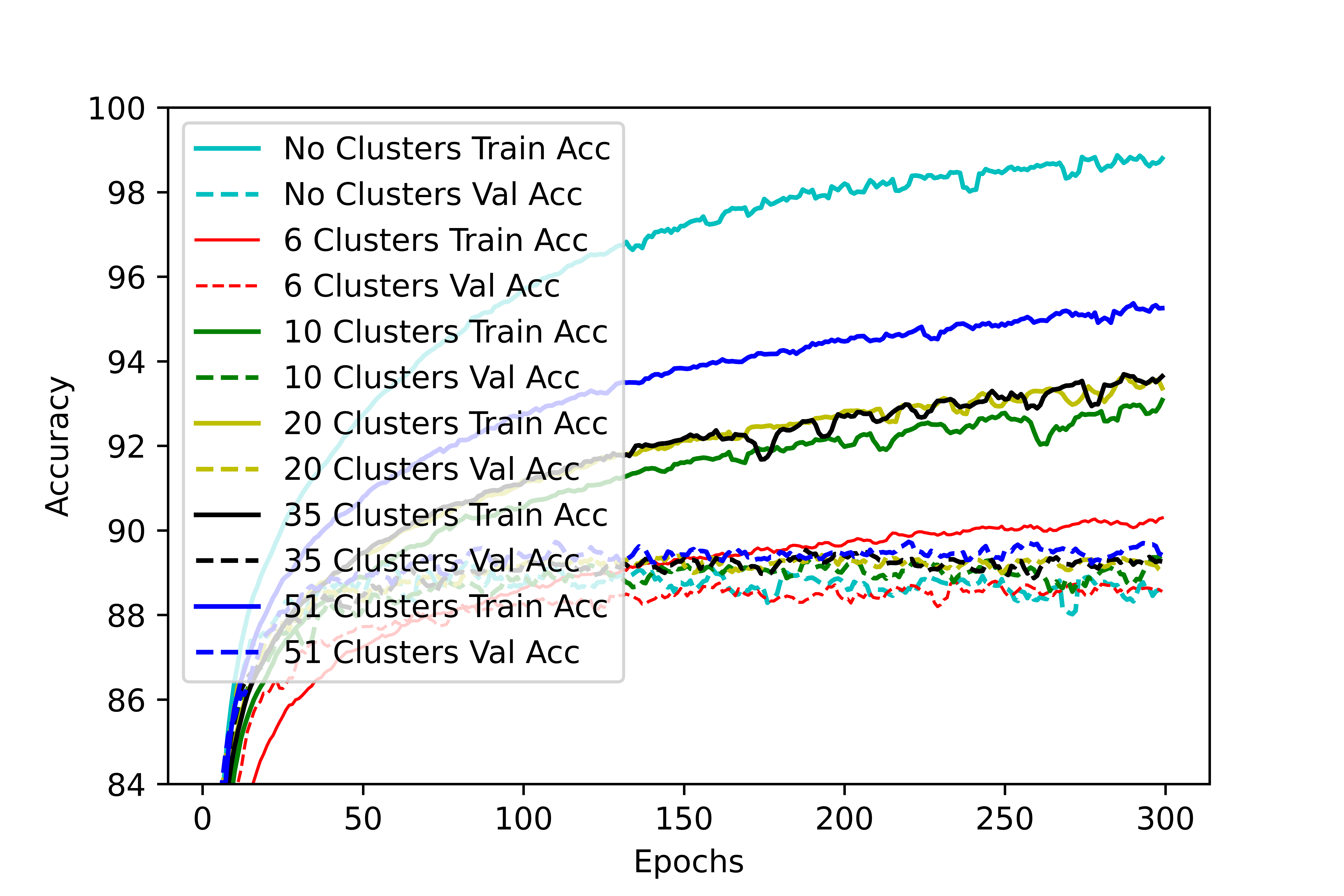}
    \includegraphics[width=0.49\linewidth]{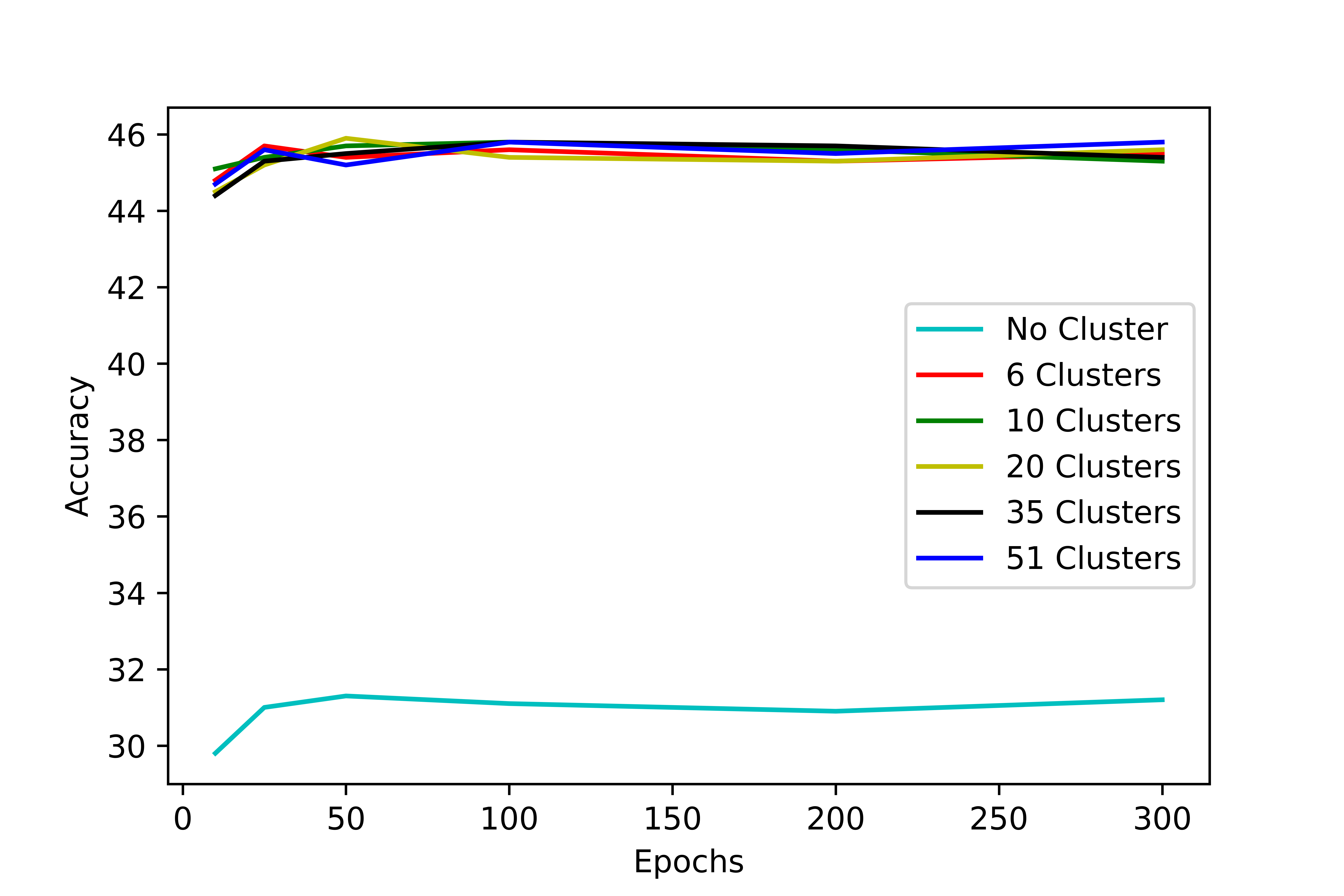}
    \caption{Left: Learning curve for the seen classes. Right: Accuracy curve for the unseen classes. The clustering-based representation avoids overfitting, which in the case of seen classes means that the gap between validation and training accuracy is smaller than in the vanilla representation. This regularization effect improves the accuracy in unseen classes. }
    \label{fig:seen_vs_unseen}
\end{figure}

\begin{figure}
    \centering
    \includegraphics[width=0.49\linewidth]{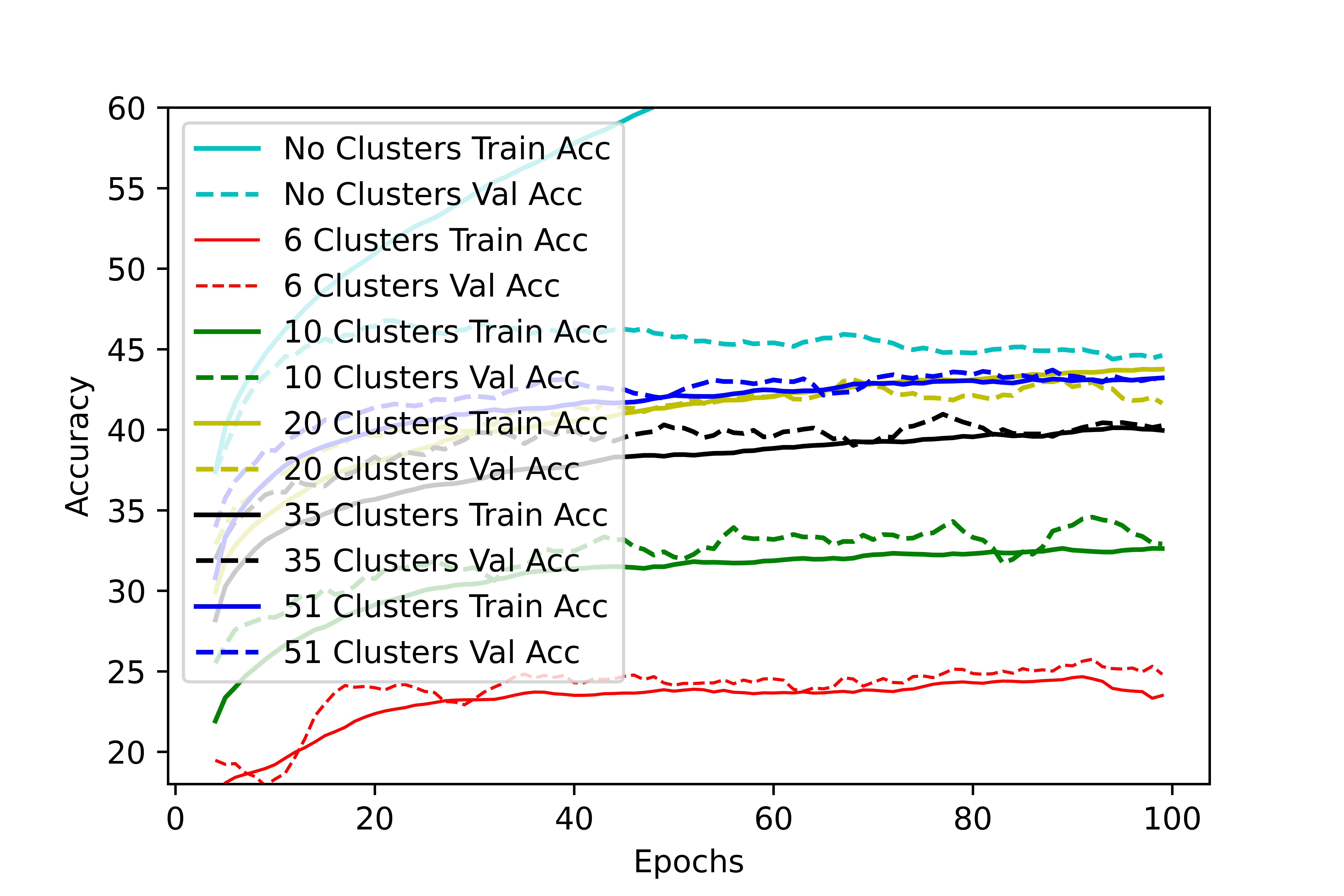}
    \includegraphics[width=0.49\linewidth]{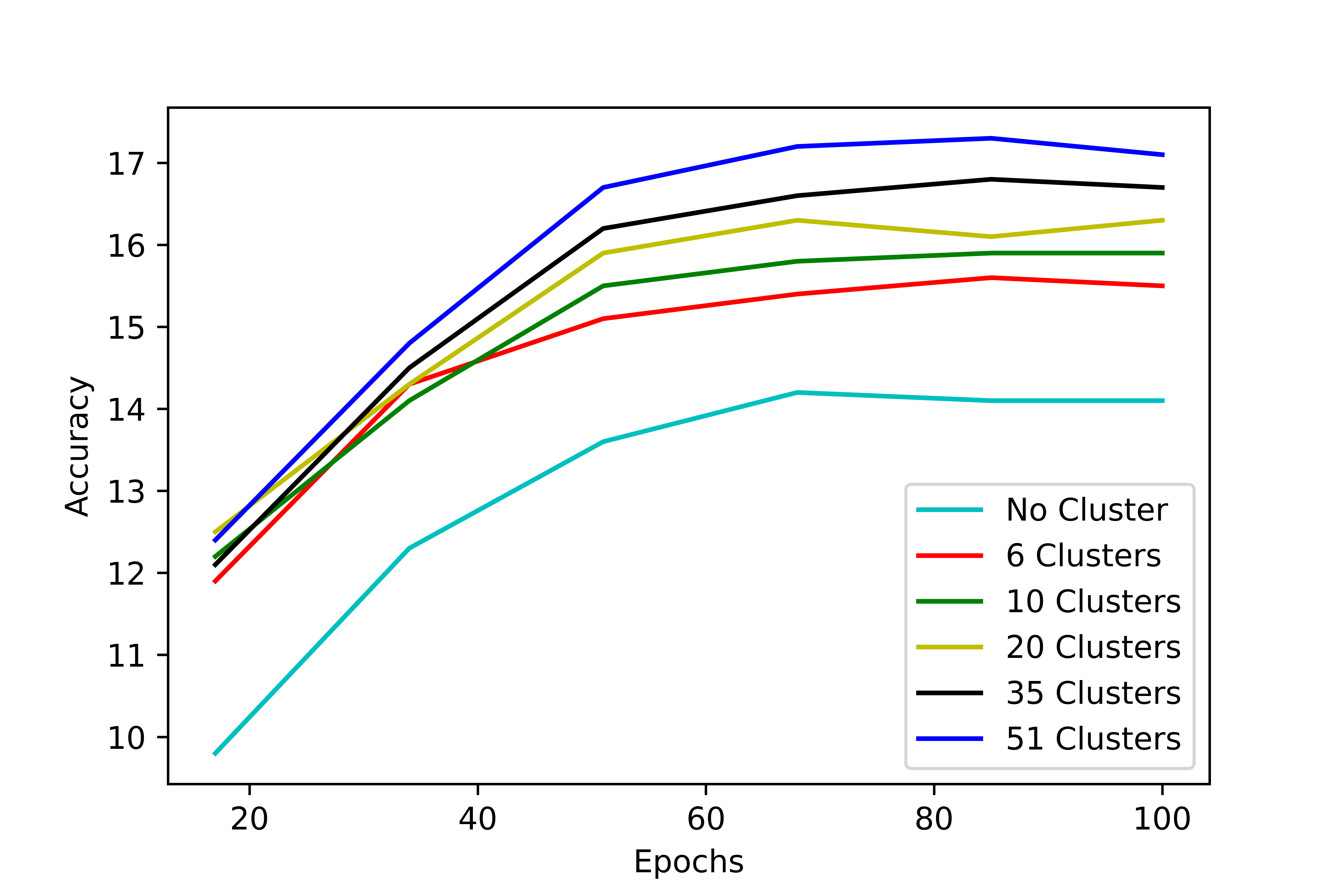}
    \caption{Left: Learning curve for the seen classes using 35\% of the data. Right: Learning curve for the unseen classes. The clustering-based representation avoids overfitting, which in the case of seen classes means that the gap between validation and training accuracy is smaller than in the vanilla representation. This regularization effect improves the accuracy in unseen classes. }
    \label{fig:seen_vs_unseen_35}
\end{figure}

In addition, we show that using only 35\% of the data of seen classes for training also benefits from clustering on the unseen classes. This can be seen in Figure~\ref{fig:seen_vs_unseen_35} While in the case of seen classes, using no clustering has the highest validation accuracy, at test time for the unseen classes, clustering leads to the best results. There are a few interesting points to note here. First, no clustering results in clear overfitting. The training accuracy reaches over 80\% while the validation accuracy reaches a peak of 46\% before dropping. However, using clustering results in the training and validation curves to be really close to each other. Another interesting point is that when there is a limited number of samples, having more clusters results in better performance at test time. This was not the case when we had all samples for the seen classes. When having all samples at training time, the number of clusters resulted in the same average accuracy as can be seen in Section~\ref{sec:nClusters}.

\section{Statistical Significance}
\label{sec:significance}

We consider the dependent t-test for paired samples. This test is utilized in the case of dependent samples, in our case different model performances on the same random data split. This is a case of a paired difference test. This is calculated as shown in Eq~\ref{ttest}.

\begin{equation}
    t=\frac{\bar{X}_{D}-\mu _{0}}{s_{D}/\sqrt{n}}
    \label{ttest}
\end{equation}

Where $\bar{X}_{D}$ is the average of the difference between all pairs and $s_{D}$ is the standard deviation of the difference between all pairs. The constant $\mu _{0}$ is zero in case we wish to test if the average of the difference is different; $n$ represents the number of samples, $n=10$ in our case. 
The comparisons can be seen in Table~\ref{tab:ttest}. The lower the value of 'p', higher the significance.

As we can see, our results are statistically significant in comparison to both OD~\cite{OD} and WGAN~\cite{clswgan} in both ZSL and GZSL. We also see that our results are statistically significant for both HMDB51 and Olympics in comparison to E2E~\cite{e2e}. In GZSL, OD~\cite{OD} also achieves results that are significantly different in comparison to WGAN~\cite{clswgan}.

\begin{table*}
\begin{center}
\begin{tabular}{|l|c|c|c|c|}
\hline
Pairs & Dataset & t-value & Statistical significance(p$<$0.05) & Type\\
\hline\hline
CLASTER and OD~\cite{OD} & UCF101 & -15.77 & Significant, p$<$0.00001 & ZSL\\
CLASTER and WGAN~\cite{clswgan} & UCF101 & -9.08 & Significant, p$<$0.00001 & ZSL\\
CLASTER and E2E~\cite{e2e} & UCF101 & -0.67 & Not Significant, p = 0.26 & ZSL\\
OD~\cite{OD} and WGAN~\cite{clswgan} & UCF101 & -1.70 & Not Significant, p$=$0.12278 & ZSL\\
\hline
CLASTER and OD~\cite{OD} & HMDB51 & -4.33 & Significant, p$=$0.00189 & ZSL\\
CLASTER and WGAN~\cite{clswgan} & HMDB51 & -5.54 & Significant, p$=$0.00036 & ZSL\\
CLASTER and E2E~\cite{e2e} & HMDB51 & -3.77 & Significant, p = 0.00219 & ZSL\\
OD~\cite{OD} and WGAN~\cite{clswgan} & HMDB51 & -3.71 & Significant, p$=$0.00483 & ZSL\\
\hline
CLASTER and OD~\cite{OD} & Olympics & -9.06 & Significant, p$<$0.00001 & ZSL\\
CLASTER and WGAN~\cite{clswgan} & Olympics & -11.73 & Significant, p$<$0.00001 & ZSL\\
CLASTER and E2E~\cite{e2e} & Olympics & -2.72 & Significant, p = 0.012 & ZSL\\
OD~\cite{OD} and WGAN~\cite{clswgan} & Olympics & -2.47 & Significant, p$=$0.03547 & ZSL\\
\hline
CLASTER and OD~\cite{OD} & UCF101 & -4.51 & Significant, p$=$0.00148 & GZSL\\
CLASTER and WGAN~\cite{clswgan} & UCF101 & -5.49 & Significant, p$=$0.00039 & GZSL\\
OD~\cite{OD} and WGAN~\cite{clswgan} & UCF101 & -3.16 & Significant, p$=$0.01144 & GZSL\\
\hline
CLASTER and OD~\cite{OD} & HMDB51 & -5.08 & Significant, p$=$0.00066 & GZSL\\
CLASTER and WGAN~\cite{clswgan} & HMDB51 & -7.51 & Significant, p$=$0.00004 & GZSL\\
OD~\cite{OD} and WGAN~\cite{clswgan} & HMDB51 & -5.27 & Significant, p$=$0.00051 & GZSL\\
\hline
CLASTER and OD~\cite{OD} & Olympics & -5.79 & Significant, p$=$0.00026 & GZSL\\
CLASTER and WGAN~\cite{clswgan} & Olympics & -8.39 & Significant, p$=$0.00002 & GZSL\\
OD~\cite{OD} and WGAN~\cite{clswgan} & Olympics & -6.22 & Significant, p$=$0.00014 & GZSL\\
\hline
\end{tabular}
\end{center}
\caption{Comparison of the t-test for different pairs of models on the same random split. Lower the value of 'p', higher the significance. As we can see, our results are statistically significant in comparison to both OD~\cite{OD} and WGAN~\cite{clswgan} in both ZSL and GZSL. For GZSL, OD~\cite{OD} also achieves results that are significant in comparison to WGAN~\cite{clswgan}. } 
\label{tab:ttest}
\end{table*}  

\section{Average of Differences in Performance for Same Splits}
\label{sec:avg_diff}

Since the performance of the model varies for each random split (as witnessed by the standard deviation values), we average the difference in performance between CLASTER, OD, WGAN and E2E on the same splits. We believe that this gives us a better metric to check the performance of CLASTER with the other approaches. The results are depicted in Table~\ref{tab:avg_diff}.

\setlength{\tabcolsep}{1.2 pt}
\begin{table}[htb]
\begin{center}
\begin{tabular}{|c|c|c|c|c| }
\hline
Models & Setting & Olympics & HMDB51 & UCF101\\
\hline\hline
Ours and WGAN \cite{clswgan} & ZSL & 17.5 $\pm$ 4.5 & 7.0 $\pm$3.8  & 17.4 $\pm$ 5.7\\
Ours and OD \cite{OD} & ZSL & 13.6 $\pm$ 4.5 & 2.4 $\pm$ 1.6 & 14.3 $\pm$ 2.7\\
Ours and E2E \cite{e2e} & ZSL & 2.6 $\pm$ 2.8 & 3.7 $\pm$ 2.8 & 0.4 $\pm$ 1.8\\
\hline
Ours and WGAN \cite{clswgan} & GZSL & 11.2 $\pm$ 4.0 & 9.3 $\pm$ 3.7 & 8.1 $\pm$ 4.4 \\
Ours and OD \cite{OD} & GZSL & 4.6 $\pm$ 2.4 & 5.2 $\pm$ 3.1 & 2.7 $\pm$ 1.8 \\
\hline
\end{tabular}
\end{center}
\caption{Comparing the average of the difference in performance for recent state-of-the-art approaches in zero-shot and generalized zero-shot action recognition on the same splits. All results were computed using sen2vec as the embedding. We can see that we outperform recent approaches in every scenario.} 
\label{tab:avg_diff}
\end{table}

\section{Number of Clusters}
\label{sec:nClusters}

\begin{figure}[htb]
    \centering
    \includegraphics[width=0.5\linewidth]{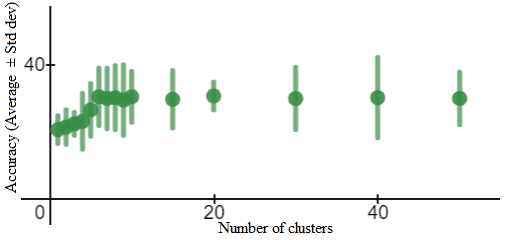}
    \caption{Effect of using different number of clusters. The green line represents the standard deviation. The reported accuracy is on the UCF101 dataset. As can be seen, the average cluster accuracy increases till about 6 clusters and then remains more or less constant. The vertical lines correspond to the standard deviation.}
    \label{fig:cluster_number}
\end{figure}

We test using different number of clusters on the UCF-101 dataset and show the results in Figure~\ref{fig:cluster_number}. These are for 5 runs on random splits. As we can see, the average accuracy increases until 6 clusters, and after that remains more or less constant. Thus, we use 6 clusters and continue with the same number for both HMDB51 and Olympics. For images, similarly, we used 5 random splits of CUB and found the performance stabilizes after having 9 clusters and use the same number of clusters for the other image datasets.

\section{Comparison of aggregation strategies and interaction between visual and semantic features}
\label{sec:ablation}

We compare the method with and without semantic features in Table 1 of the main paper. Below, in Table~\ref{tab:multimodalinteraction} we show other aggregation options such as averaging and dot product. All results are using ED as semantic embedding.  
\begin{table}[htb]
\centering
\begin{tabular}{|l|l|}
\hline
\multicolumn{1}{|l|}{Method} & \multicolumn{1}{l|}{HMDB51} \\ \hline
Average                      &       33.1 $\pm$ 2.9                      \\
Dot Product                  &       33.9 $\pm$ 3.2                      \\
Weighted Average             &       35.3 $\pm$ 3.6                      \\
Concatenation                &       43.2 $\pm$ 1.9              \\
\hline
\end{tabular}
\caption{Results on different aggregation options for the semantic and visual embeddings.}
\label{tab:multimodalinteraction}
\end{table}

\section{Seen and Unseen Class Performance for GZSL}
\label{sec:gzsl:seenunseen}

In order to better analyze performance of the model on GZSL, we report the average seen and unseen accuracies along with their harmonic mean. The results using different embeddings and on the UCF101, HMDB51 and Olympics datasets are reported in Table~\ref{tab:gzsl_vid}. The reported results are on the same splits for fair comparison~\cite{truze}.

\setlength{\tabcolsep}{1.3pt}
\begin{table}[htb]
\begin{center}
\begin{tabular}{| *{11}{c|} }
\hline
Model & E & \multicolumn{3}{c|}{Olympics} & \multicolumn{3}{c|}{HMDB51}& \multicolumn{3}{c|}{UCF-101}\\
\hline
& & u & s & H & u & s & H & u & s & H \\
\hline\hline
WGAN \cite{clswgan} & A & 50.8 & 71.4 & 59.4 & - & - & - & 30.4 & 83.6 & 44.6\\
OD \cite{OD} & A & 61.8 & 71.1 & 66.1 & - & - & - & 36.2 & 76.1 & 49.1 \\
CLASTER & A & 66.2 & 71.7 & 68.8 & - & - & - & 40.2 & 69.4 & 50.9 \\
\hline
WGAN \cite{clswgan} & W & 35.4 & 65.6 & 46.0 & 23.1 & 55.1 & 32.5 & 20.6 & 73.9 & 32.2\\
OD \cite{OD} & W & 41.3 & 72.5 & 52.6 & 25.9 & 55.8 & 35.4 & 25.3 & 74.1 & 37.7\\
CLASTER & W & 49.2 & 71.1 & 58.1 & 35.5 & 52.8 & 42.4 & 30.4 & 68.9 & 42.1 \\
\hline
WGAN \cite{clswgan} & S & 36.1 & 66.2 & 46.7 & 28.6 & 57.8 & 38.2 & 27.5 & 74.7 & 40.2\\
OD \cite{OD} & S & 42.9 & 73.5 & 54.1 & 33.4 & 57.8 & 42.3 & 32.7 & 75.9 & 45.7 \\
CLASTER & S & 49.9 & 71.3 & 58.7 & 42.7 & 53.2 & 47.4 & 36.9 & 69.8 & 48.3\\
\hline
CLASTER & C & 66.8 & 71.6 & 69.1 & 43.7 & 53.3 & 48.0 & 40.8 & 69.3 & 51.3\\
\hline
\end{tabular}
\end{center}
\caption{Seen and unseen accuracies for CLASTER on different datasets using different embeddings. 'E' corresponds to the type of embedding used, wherein 'A', 'W', 'S' and 'C' refers to manual annotations, word2vec, sen2vec and combination of the embeddings respectively. 'u', 's' and 'H' corresponds to average unseen accuracy, average seen accuracy and the harmonic mean of the two. All the reported results are on the same splits.} 
\label{tab:gzsl_vid}
\end{table}

\section{Conclusion}

Zero-Shot action recognition is the task of recognizing action classes without any visual examples. The challenge is to map the knowledge of seen classes at training time to that of novel unseen classes at test time. We propose a novel model that learns clustering-based representation of visual-semantic features, optimized with RL. We observe that all three of these components are essential. The clustering helps regularizing, and avoids overfitting to the seen classes. The visual-semantic representation helps improve the representation. And the RL yields better, cleaner clusters. 
The results is remarkable improvements across datasets and tasks over all previous state-of-the-art, up to 11.9\% absolute improvement on HMDB51 for GZSL. 

\clearpage

\bibliographystyle{splncs04}
\bibliography{eccv}
\end{document}